\pdfoutput=1

\documentclass[11pt]{article}

\usepackage{acl}

\usepackage{times}
\usepackage{latexsym}
\usepackage{graphicx}
\usepackage{multirow}

\usepackage{amsfonts}

\usepackage[T1]{fontenc}

\usepackage[utf8]{inputenc}

\usepackage{microtype}

\title{Semantically Consistent Data Augmentation for Neural Machine Translation via Conditional Masked Language Model}

\author{Qiao Cheng, Jin Huang, Yitao Duan \\
        NetEase Youdao \\ 
        Beijing, China \\
        \texttt{\{chengqiao, huangjin, duan\}@rd.netease.com}}

\begin{document}
\maketitle
\begin{abstract}

This paper introduces a new data augmentation method for neural machine translation that can enforce stronger semantic consistency both within and across languages. Our method is based on Conditional Masked Language Model (CMLM) which is bi-directional and can be conditional on both left and right context, as well as the label. We demonstrate that CMLM is a good technique for generating context-dependent word distributions. In particular, we show that CMLM is capable of enforcing semantic consistency by conditioning on \textit{both} source and target during substitution. In addition, to enhance diversity, we incorporate the idea of \textit{soft} word substitution for data augmentation which replaces a word with a probabilistic distribution over the vocabulary. Experiments on four translation datasets of different scales show that the overall solution results in more realistic data augmentation and better translation quality. Our approach consistently achieves the best performance in comparison with strong and recent works and yields improvements of up to 1.90 BLEU points over the baseline. \footnote{Our code is available at \url{https://github.com/netease-youdao/cmlm\_da}.}

\end{abstract}

\section{Introduction}
Neural network models have achieved remarkable results in many fields such as computer vision, natural language processing, and speech. In order to obtain adequate expressivity, the models usually come with a large number of parameters. However, such models are prone to overfitting if trained with an insufficient amount of training data. Data Augmentation (DA) is an effective technique that has been used in many areas to augment existing labeled data and boost the performance of machine learning models. For example, in computer vision, training data is often augmented by ways such as horizontal flipping, random cropping, tilting, and color shifting \citep{krizhevsky2012imagenet, cubuk2018autoaugment}. While DA has become a standard technique to train deep networks for image processing, it is relatively under-explored in Natural Language Processing (NLP).



The exact mechanisms and theoretical foundations of data augmentation are still under investigation. 
Most studies show empirically that data augmentation is effective and provide some intuitive explanations. A recent work in the field of vision \citep{gontijo2020affinity} demonstrates that affinity (the distributional shift caused by DA) and diversity (the complexity of the augmentation) can predict the performance of data augmentation methods. However, neither metric can be measured without completing the entire DA process. Therefore, it is still challenging to evaluate the goodness of a DA technique without full-fledged experimentation and it is not clear how the result can be used to guide the design of data augmentation schemes.

Generally speaking, DA can be classified into two categories. The first tries to produce realistic samples that resemble the inherent semantics of naturally generated data.  In areas such as computer vision, this is often achieved via heuristics that mimic the intrinsic processes that could have actually happened in the physical world, such as photometric noise, flipping, and scaling, etc. The second perturbs the data in a stochastic fashion, resulting in unrealistic samples. Some (e.g., \citet{Bishop95noise}) interpret this as a type of regularization that boosts model performance by reducing overfitting. Both are being exploited in NLP.




This paper focuses on lexical replacement methods that augment the training data by altering existing sentences in the parallel corpus of a neural machine translation (NMT) system. We have observed frequently in practice, as well as in literature \citep{gao2019soft,fadaee2017data,kobayashi2018contextual,wu2019conditional,dong2021data,liu2021counterfactual}, that augmented data samples that preserve the semantics of the real labeled data increase the effective training size and are beneficial for model performance. We call this property \textit{semantic consistency}. In the case of NMT, the training data comes in the form of a collection of \texttt{<source, target>} sentence pairs where \texttt{source} is a sentence in the source language and \texttt{target} its translation in the target language. Semantic consistency requires that (1) both \texttt{source} and \texttt{target} are fluent and grammatically correct in their respective languages; and (2) \texttt{target} is a high quality translation of \texttt{source}.

\begin{table}[h]
\begin{tabular}{p{0.1\textwidth}p{0.3\textwidth}}
\hline
\textbf{German}    & Es ist ja ganz \textbf{angenehm}, in eine kleine Klasse zu kommen. 
  \\ \hline
\textbf{English}    & You know, it's very \textbf{pleasant} to walk into a small class.
 \\ \hline
\textbf{Case 1}    & You know, it's very \textbf{please} to walk into a small class. \\
\textbf{Case 2}    & You know, it's very \textbf{uncomfortable} to walk into a small class. \\ \hline
\textbf{Case 3}    & You know, it's very \textbf{enjoyable/comfortable} to walk into a small class. \\ \hline
\end{tabular}
\caption{Data augmentation examples with varying degrees of semantic consistency.}
\label{table:example}
\end{table}

Existing methods augment the training data using word swapping, removal or substitution \citep{artetxe2017unsupervised, lample2017unsupervised} on either \texttt{source} or \texttt{target}, or both.  Due to the discrete nature of language, these transformations are not always semantic-preserving. Quite often they either weaken the fluency of \texttt{source} or/and \texttt{target}, or break their relationships. 
To illustrate, consider the example given in Table \ref{table:example} that shows a sentence pair from an English-German parallel corpus. Case 1 to 3 are three synthetic English sentences generated by some DA processes. Both Case 1 and 2 are undesirable because the former, although substituting the word \textbf{pleasant} with a word close in meaning, is grammatically incorrect, whereas the latter is not a good translation of the German sentence. Case 3, on the other hand, is a good augmentation that satisfies the two requirements of semantic consistency.



 
\subsection{Our Contributions} To achieve better augmentation, the generation process must make better use of context and label. In this paper, we introduce Conditional Masked Language Model (CMLM) \citep{wu2019conditional,ghazvininejad-etal-2019-mask,chen2020distilling} to data augmentation for NMT. A Masked Language Model can make use of both left and right context, and a CMLM is an enhanced version that can be conditional on more information.  CMLM has been used successfully in tasks such as text classification \citep{wu2019conditional}. However, to the best of our knowledge, its application to text generation, especially using deep bidirectional models such as BERT \citep{devlin-etal-2019-bert}, has not been explored. We demonstrate in this paper that CMLM is a good technique for generating context-dependent word distributions. In particular, we show that CMLM is capable of enforcing semantic consistency by conditioning on \textit{both} source and target during substitution. In addition, to enhance diversity, we combine the \textit{soft} word substitution approach for DA, which replaces a word with a probabilistic distribution over the vocabulary \citep{gao2019soft}. Experiments on four translation datasets of different scales show that the overall solution results in more realistic data augmentation and better translation quality. Our approach consistently achieves the best performance in comparison with strong and recent works and yields improvements of up to 1.90 BLEU points over the baseline.

In addition, we introduce an unsupervised method to measure semantic consistency without full-fledged training of NMT models, which may take many days even on GPU clusters. This could be used to provide an efficient early assessment of a data augmentation scheme. 


\section{Related Work}


From a technical perspective, previous work on data augmentation for NLP can be classified as either context-independent or context-dependent. Context-independent approaches often apply predetermined, easy-to-compute transformations that depend solely on the word or sentence to be altered. Not surprisingly, most of them are not semantically consistent.
\citet{wei2019eda} improves performance on many text classification tasks through a set of word level random perturbation operations, including random insertion, deletion, and swapping. Similar ideas have been applied to NMT, but the methods differ in how and what to alter. \textit{Swap} \citep{artetxe2017unsupervised, lample2017unsupervised} randomly swaps words in nearby positions within a window size \textit{k} and \textit{Drop} \citep{iyyer2015deep, lample2017unsupervised} randomly drops word tokens. \textit{Blank} \citep{xie2017data} replaces the candidate word with a placeholder token and \textit{Smooth} \citep{xie2017data} replaces it with a word sampled from the frequency distribution of the vocabulary, showing that data noising is an effective regularizer for NMT.
\textit{SwitchOut}, introduced in \cite{wang2018switchout}, formulates the design of a DA algorithm as an optimization problem that maximizes an objective that encourages two desired properties: smoothness and diversity. \textit{SwitchOut} independently replaces words in both \texttt{source} and \texttt{target} by other words uniformly sampled from their respective vocabularies.  Others try to preserve a certain level of semantic consistency by replacing words with their synonyms selected from a handcrafted ontology such as WordNet \citep{zhang2015character} or words based on similarity calculation \citep{wang2015s}.

These works do not make use of important context and label information and, in practice, usually cause a very small or even negative impact on performance. Context-dependent approaches, on the other hand, modify words, phrases, or the whole sentence based on their contextual information that is usually modeled using neural networks. We summarize a few representative ones below.

\citet{fadaee2017data} propose a simple but effective approach to augment the training data of NMT for low-resource language pairs. Their work uses shallow LSTM language models (LM) trained on large amounts of monolingual data to first substitute a word in \texttt{source}, and then put the corresponding translation in \texttt{target}, using automatic word alignments and the traditional statistical MT practice. \textit{LM$_{sample}$} \citep{kobayashi2018contextual} proposes contextual augmentation for text classification by offering a wide range of substitute words, which are predicted by a label-conditional bidirectional language model.  \citet{wu2019conditional} retrofit BERT to conditional BERT that allows it to augment sentences without breaking the label-compatibility. The BERT-based solution brings two benefits. First, BERT's Transformer core provides a more structured memory for handling long-term dependencies in text. Second, BERT, as a deep bidirectional model, is strictly more powerful than the shallow concatenation of left-to-right and right-to-left models.

A recent work \citep{liu2021counterfactual} treats a translation language model as a causal model and performs data augmentation by counterfactual-based causal inference. Their DA replaces source phrases according to a masked language model and the aligned target phrase by a cross-lingual language model (XLM) \citep{conneau2019cross} conditional on the changed source phrase.

Different from their work, we use two separate CMLMs to augment source and target respectively, which means that, instead of model prediction, the condition is always true information for both CMLMs. We show its superiority in section \ref{sec.sen.consistency}.

The way we incorporate augmented data into the NMT training is drawn from the idea of ``soft'' word introduced by \textit{SCA} \citep{gao2019soft}. Basically, the embedding of a chosen word in a sentence is replaced by its probabilistic distribution predicted by a language model. This brings in more diversity to the DA process. However, \citet{gao2019soft} as a DA solution is based on a uni-directional language model and is \textit{not} label-conditional. As we show in section \ref{sec.mainresult} that this is less optimal.






\begin{figure*}[htb]
  \centering
  \includegraphics[scale=0.45,width=0.78\textwidth]{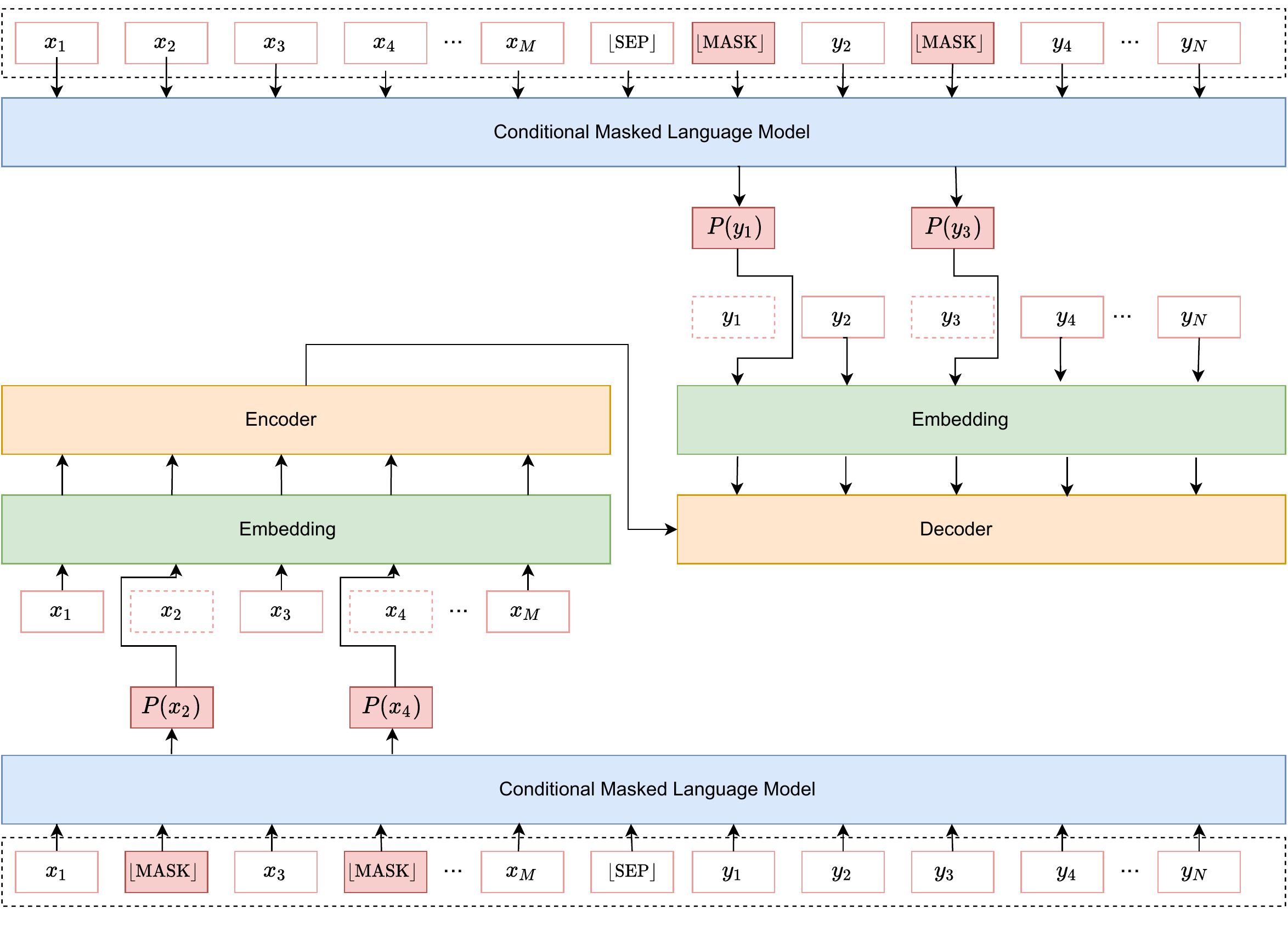}
  \caption{The architecture of our CMLM-based soft contextual data augmentation approach. }
  \label{fig:traing_arch}
\end{figure*}

\section{Approach}
In this section, we present our method in detail. We first introduce conditional MLM, then we show how to apply CMLM to data augmentation in neural machine translation tasks.

Let $(X, Y)$ be a pair of source and target sentences where $X = (x_1, x_2, \ldots, x_M)$ and $Y = (y_1, y_2, \ldots, y_N)$ are two sequences of tokens in source and target languages, with lengths $M$ and $N$, respectively. A neural machine translation system learns the conditional probability $p(y_1, y_2, \ldots, y_N | x_1, x_2, \ldots, x_M)$.






\subsection{Conditional MLM}
\label{sec.cmlm}

Recall that our goal is to augment NMT's parallel corpus with synthesized data that preserves the semantics within source and target sentences, as well as their cross-lingual relations. To this end, we resort to Conditional MLM for generating context-dependent word distributions, with which we then find the best substitutes for a given word.  CMLM is a variation of MLM, which allows further fine-tuning of the pre-trained model. It makes the strong assumption that the masked tokens are conditionally independent of each other given the context and predicts the probabilities individually \cite{ghazvininejad-etal-2019-mask}. 

In our case, we apply the following two practices when instantiating our CMLMs:

\begin{itemize}
\item We condition the CMLM on \textit{both} $X$ and $Y$. 

\item During the training of a CMLM, we only mask out tokens in either $X$, or $Y$, but not both.  
\end{itemize}

We call this approach ``\textbf{Conditioning on Both but Predicting One}'', referring to how it treats the source and target sides in the NMT training. Specifically, for each sentence pair $(X, Y)$,  we first concatenate $X$ and $Y$, then randomly mask 15\% of the words in $X$, and then train a CMLM to predict the masked words:

\begin{equation} \label{eqn.maskx}
P\left(x_{1}^{m}, \ldots, x_{i}^{m} \mid X^{u}, Y\right)
\end{equation}

where $x_{i}^m$ denotes a masked token and $X^u$ the unmasked ones within $X$. For the tokens in the target sentence, we train a \textit{separate} CMLM to get their distribution similarly:

\begin{equation} \label{eqn.masky}
P\left(y_{1}^{m}, \ldots, y_{i}^{m} \mid X, Y^{u}\right)
\end{equation}

During the training of an NMT model, both $X$ and $Y$ are available. Conditioning on the reference sentence $Y$ allows the model to enforce stronger consistency between input and label, resulting in meaningful translations when applied to DA in NMT. We show in section \ref{sec.sen.consistency}, using metrics developed for translation quality estimation, that this choice significantly improves the translation quality of the generated sentence pairs.

Changing $X$ or $Y$ but not both for DA is a deliberate choice. Typical modern languages have diverse vocabularies, with synonyms and semantically equivalent or close expressions. This already provides abundant opportunities for semantic-preserving transformations. Therefore, it is not necessary to alter $X$ and $Y$ simultaneously. In section \ref{sec.sen.consistency}, we compare our choice with an XLM (cross-lingual language model) \citep{conneau2019cross} approach which changes $X$ and $Y$ simultaneously. The empirical study shows that our approach can avoid introducing incorrect \texttt{<source, target>} pairs and improve NMT performance.

\subsection{Soft Conditional Contextual DA}
\label{sec.soft}

Once a CMLM is trained, one could use it to expand training data for NMT. This is typically done by replacing words with others predicted by the language model at the corresponding positions (e.g., \citet{kobayashi2018contextual,wu2019conditional}). In our case, since the probability distribution of the masked words $P\left(x_{1}^{m}, \ldots, x_{i}^{m} \mid X^u, Y\right)$, or $P\left(y_{1}^{m}, \ldots, y_{i}^{m} \mid X, Y^{u}\right)$ if we mask out words in $Y$, contains information from both backward and forward contexts, as well as target sentence, sampling from such distribution could potentially generate better substitutions for the word on the masked position. However, such a method could be expensive: to generate enough samples with adequate variation, exponentially many candidates have to be processed. 

Instead, inspired by \citet{gao2019soft}, we take a \textit{soft} approach. In essence, this method works directly with the word embeddings and uses the expectation of a word's embedding over the CMLM's output distribution to replace its original embedding. Let $w$ be a candidate word and $P(w)$ its distribution defined by the CMLM. Note that $P(w)$ is conditional on the context that we described earlier and is over the entire vocabulary. Suppose $E$ is the embedding matrix of all the $|V|$ words. We use $E_W$ to denote the embedding vector of a word $W$. The embedding of the soft word $w$ is:
\begin{equation} \label{eqn.soft}
e_{w}=\mathbb{E}_{W \sim P(w)}[E_W] =\sum_{j=0}^{|V|} p_{j}(w) E_{j}
\end{equation}

\subsection{NMT Training with DA}
\label{sec.training}

In this section, we elaborate on the training process of the NMT model with our DA method. Figure \ref{fig:traing_arch} shows the architecture of the scheme. There are two independently trained CMLMs, one for augmenting the encoder, and the other the decoder. The two CMLMs can be turned on/off independently and we study the effects in section \ref{sec.enc_dec}.

We use BERT \cite{devlin-etal-2019-bert} as our CMLM, for its deep bidirectional natural, and superior capability for handling long-term dependencies. We start by taking a pre-trained multilingual BERT, and fine-tune it using the method described in \ref{sec.cmlm}. The NMT training proceeds as usual, except that, at each sentence pair $(X,Y)$, for each word in $X$ (or $Y$), with probability $\gamma$ we replace its embedding by its soft version defined by Equation \ref{eqn.soft}. Notice that, our method does \textit{not} generate any data explicitly. Rather, we use embedding substitution to incorporate augmentation directly into the training process. We study the effect of different values of  $\gamma$ in section \ref{sec.gamma}.

\section{Experiments}
In this section, we demonstrate the effectiveness of our method on four datasets with diverse language variation. They include three relatively small-scale datasets, \{German, Spanish, Hebrew\} to English (\{De, Es, He\}-> En) from the well-known IWSLT 2014, and one large-scale English to German (En->De) dataset from WMT14 . 

\subsection{Data}
For IWSLT14 De->En task we follow the same pre-processing steps and the same train/dev/test split as in \citet{gao2019soft}. The training dataset and validation dataset contains about 160K and 7K sentence pairs, respectively. Consistent with previous work, tst2010, tst2011, tst2012, dev2010, and dev2012 are concatenated as our test data. For Es->En and He->En tasks, there are 181K and 151K parallel sentence pairs in each training set. We validate on tst2013 and test on tst2014 for these two tasks. For all IWSLT translation tasks, we use a joint source and target vocabulary with 10K byte-pair-encoding (BPE) \citep{sennrich2016neural} types.
For the WMT2014 En-De translation task, again, we follow \citet{gao2019soft} to filter out 4.5M sentence pairs for training. We concatenate newstest2012 and newstest2013 as the validation set and use newstest2014 as the test set. We use a joint source and target vocabulary built upon the BPE with 40k sub-word types. For fair comparison to previous work, we report tokenized BLEU \citep{papineni2002bleu} scores computed with the \textit{multi-bleu.perl} script from Moses.\footnote{https://github.com/moses-smt/mosesdecoder/blob/master/scripts/generic/multi-bleu.perl} To further boost comparability, we also report detokenized BLEU scores computed using sacreBLEU \citep{post-2018-call}. \citep{post-2018-call}. For all experiments, we performed significance tests based on bootstrap resampling introduced by \citet{koehn2004statistical}. 

\begin{table*}[h!]
\centering
\begin{tabular}{llllll}
\hline
\multirow{2}{*}{}    & \multicolumn{3}{c}{\textbf{IWSLT}}                                          &                      & \textbf{WMT}                  \\ \cline{2-4} \cline{6-6} 
                     & \textbf{De->En}                   & \textbf{Es->En}                   & \textbf{He->En}                   &                      & \textbf{En->De}                   \\ \hline
\multicolumn{6}{c}{Other Reported Results}   \\ \hline
Base$^\ast$                 & 34.79                    & 41.58                    & 33.64                    &                      & 28.40                    \\ 
\textit{Swap}$^\ast$ & 34.70 & 41.60 & 34.25 &   & 28.13 \\ 
\textit{Drop}$^\ast$  & 35.13 & 41.62 & 34.29 &   & 28.29 \\ 
\textit{Blank}$^\ast$  & 35.37 & 42.28 & 34.37 &   & 28.89 \\ 
\textit{Smooth}$^\ast$  & 35.45 & 41.69 & 34.61 &   & 28.97 \\ 
\textit{LM$_{sample}$}$^\ast$  & 35.40 & 42.09 & 34.31 &   & 28.73 \\ 
\textit{SCA}$^\ast$  & \textbf{35.78} & \textbf{42.61} & \textbf{34.91} &  & \textbf{29.70} \\ 
mixSeq$^\dagger$  & \textbf{35.78} & 41.39 & - &  & 29.61 \\ \hline
\multicolumn{6}{c}{Our Implementations}   \\ \hline
Base                 & 34.37                    & 41.67                    & 33.76                    &                      & 28.25                    \\ 
CMLM$_{hard}$ & 35.76 & 42.25 & 34.66 &  & 30.01 \\ 
CMLM$_{soft}$ & \textbf{35.93(+1.56)} & \textbf{42.92(+1.25)} & \textbf{35.21(+1.45)} &  & \textbf{30.15(+1.9)} \\ \hline
\end{tabular}
\caption{BLEU scores over the test sets. ($\ast$) from \citet{gao2019soft}. ($\dagger$) from \citet{wu2021mixseq} }
\label{table:1}
\end{table*}

\begin{table*}[h!]
\centering
\begin{tabular}{llllll}
\hline
\multirow{2}{*}{}    & \multicolumn{3}{c}{\textbf{IWSLT}}                                          &                      & \textbf{WMT}                  \\ \cline{2-4} \cline{6-6} 
                     & \textbf{De->En}                   & \textbf{Es->En}                   & \textbf{He->En}                   &                      & \textbf{En->De}                   \\ \hline
Base                 & 33.62                    & 40.87                    & 33.15                    &                      & 27.49                    \\ 
CMLM$_{hard}$ & 35.07 & 41.45 & 34.01 &  & 29.08 \\ 
CMLM$_{soft}$ & \textbf{35.31(+1.69)} & \textbf{42.01(+1.14)} & \textbf{34.51(+1.36)} &  & \textbf{29.37(+1.88)} \\ \hline
\end{tabular}
\caption{SacreBLEU scores over the test sets.}
\label{table:sacrebleu}
\end{table*}


\subsection{Model} 
\label{sec.model}

Our model uses the Transformer architecture, which is solely based on attention mechanisms and dominates most of the sequence-to-sequence tasks. For all IWSLT tasks, we adopt the transformer\_iwslt\_de\_en configuration for the NMT model. Specifically, both the encoder and decoder consist of 6 blocks, and the source and target word embedding are shared for the language pair. The dimensions of embedding and feed-forward sub-layer are set to 512 and 1024, respectively. The number of attention heads is set to 4. The default dropout rate is 0.3. For WMT14 En-De, we use the default transformer\_big configuration for the NMT model. Specifically, the dimensions of embedding and feed-forward sub-layer are 1024 and 4096, respectively. The NMT models are trained by Adam \citep{kingma2015adam} optimizer with default learning rate schedule as \citet{vaswani2017attention}.

For all tasks, we adopt the BERT-base configuration for the CMLM model, except that the number of hidden layers is set to 4 to speed up the training process. We use the bottom 4 layers of the pre-trained BERT-base-multilingual-cased model as the starting point of CMLM fine-tuning. We also experiment with an entirely randomly-initialized CMLM model and find that the pre-trained weights result in faster CMLM training. We follow \citet{devlin-etal-2019-bert} for the CMLM fine-tuning and use a triangular learning rate schedule with maximum learning rate $\eta$. The CMLM parameters are also updated with the Adam optimizer.

\subsection{Main Results}
\label{sec.mainresult}

We compare our method against several other strong data augmentation methods, including several context-independent approaches such as \textit{Swap} \citep{artetxe2017unsupervised, lample2017unsupervised}, \textit{Drop} \citep{iyyer2015deep, lample2017unsupervised}, \textit{Blank} \citep{xie2017data} and \textit{Smooth} \citep{xie2017data}, and two context-dependent ones, \textit{LM$_{sample}$} \citep{kobayashi2018contextual} and \textit{SCA} \citep{gao2019soft}. We also compare it against a sentence-level augmentation method, mix{S}eq \citep{wu2021mixseq}, which randomly selects two sentence pairs, concatenates the source sentences and the target sentences, respectively, with a special label \textrm<sep> separating two samples, and trains the model on such augmented dataset. 

Our baseline is the vanilla transformer described earlier without DA. For comparison, we performed two sets of data augmentation experiments using CMLM: (1) CMLM$_{soft}$ uses the soft approach described in section \ref{sec.soft} and follows the training framework in section \ref{sec.training}. (2) CMLM$_{hard}$ uses the conventional hard substitution approach, with the substitution words generated by sampling from the CMLMs. Both CMLM$_{soft}$ and CMLM$_{hard}$ augment both the encoder and the decoder, and use the same mask probability $\gamma = 0.25$, which we find to be the optimal configuration. See sections \ref{sec.enc_dec} and \ref{sec.gamma}.

The BLEU and SacreBLEU scores on four translation tasks are presented in Table \ref{table:1} and \ref{table:sacrebleu}, respectively. Both CMLM$_{soft}$ and CMLM$_{hard}$ are superior to the base system, with CMLM$_{soft}$ consistently achieves the best performance on all tasks and across all comparisons. The CMLM (soft) approach significantly outperformed the baseline in Table \ref{table:1} and Table \ref{table:sacrebleu} for all four tasks, with $p$-values lower than 0.02. Most remarkably, our DA improves the baseline by as much as 1.90 BLEU points on the WMT14 En->De dataset. 


In addition to experiments on publicly available corpora, we also evaluate the scheme on Youdao's production NMT engine, \footnote{\url{https://fanyi.youdao.com/}} a major multilingual neural machine translation service that is trained with data at least three orders of magnitudes larger than the public corpora. The method achieves similar consistent improvements. Our DA mechanism has been built into the production NMT engine, serving billions of requests each day.

\section{Analysis}
Our method consists of multiple modules, and we design several groups of comparative experiments to analyze their effects.

\subsection{Semantic Consistency} \label{sec.sen.consistency}


Recall that the ``soft'' substitution approach that we use works directly with embeddings and does not generate synthetic data explicitly. The quality of the DA process depends on the distributions defined by the two CMLMs (equations \ref{eqn.maskx} and \ref{eqn.masky}). There is no straightforward metric to measure the distributions in terms of semantic consistency. Here we propose a simple sampling-based approach. The intuition is: if the distribution is used for text generation, the quality of resulting sentence pairs is a good indicator of the effectiveness of its role in the DA process.

Specifically, given a sentence pair $(X, Y)$, we randomly replace some tokens from $X$ (resp. $Y$) with those sampled from the source (resp. target) CMLM, resulting in $(X^\prime, Y)$ (resp. $(X, Y^\prime)$). We manually inspect a small sample and find that our method indeed produces sentence pairs that are generally both fluent in their respective languages and correct in terms of translation quality. However, our goal is to have an automatic method that can be used to assess semantic consistency at large scale. To this end, we draw on the research in Quality Estimation (QE) for Machine Translation. Self-Supervised QE aims to evaluate the quality of machine-translated sentences without human labeling, which aligns perfectly with our goal.

\citet{zheng2021qeemnlp} show that the conditional probability computed by the CMLM in Equation \ref{eqn.masky} is a good indicator of translation quality (which also implies fluency). Specifically, let $y^m$ be a word in the target, the translation quality score of this word is defined as $P\left(y^{m} \mid X, Y^{u}\right)$ as computed by the CMLM. The sentence-level quality score is simply averaging the quality scores over all target words.

Our case is slightly different. Since we have both $X$ and $Y$, we can use the idea of \citet{zheng2021qeemnlp} but with a more direct approach: we can compare the words in $X^\prime$ (resp. $Y^\prime$) against the original ones in $X$ (resp. $Y$) and compute the accuracy. This is equivalent to taking expectations over the test sentences.

\begin{table}[h]
\begin{tabular}{lccc}
\hline
  &\textbf{Source Acc}  &\textbf{Target Acc}  &\textbf{BLEU} \\ \hline
MLM        & 53.5\% & 44.0\% & 35.56 \\ 
XLM        & 74.8\% & 70.4\% & 35.65 \\ 
CMLM         & 80.1\% & 75.5\% & 35.93\\ \hline
\end{tabular}
\caption{The prediction accuracy of source and target, and BLEU for IWSLT14 German-English translation.}
\label{table:sc.acc}
\end{table}

We compare our CMLM-based approach against the DA results from (1) an XLM-based scheme in \citet{liu2021counterfactual}, which alters both $X$ and $Y$ by treating a translation language model as a causal model and performing data augmentation by counterfactual-based causal inference; and (2) a simple MLM which does not condition on any portion of $Y$. All implementations use models with the same configuration as the CMLM described in section  \ref{sec.model}, fine-tuned with the same training data but their individual conditions and objectives. Table \ref{table:sc.acc} shows the prediction accuracy of masked words on the 7K IWSLT14 German-English validation data set. Consistent with the mask probability during CMLM training, we let the model predict 15\% of the words in $X$ or $Y$. For ease of comparing the final effects on the machine translation task, Table \ref{table:sc.acc} also shows the BLEU scores measured on IWSLT14 German-English dataset after applying the DA method to the NMT engine.

Our CMLM-based solution achieves strong prediction accuracy rates of 80.1\% and 75.5\% on source and target sides, respectively, significantly outperforming the MLM approach by near 30 percentage points. This shows that our method is capable of generating synthetic sentence pairs with much better translation quality. The improvement over XLM is milder but still significant, with 5+ percentage points. BLEU scores follow a similar trend. Recall that we use independent CMLMs to alter either $X$ or $Y$ but not both, while XLM uses a single cross-lingual language model to change both. The results confirm our conjecture that altering both $X$ and $Y$ simultaneously while preserving semantic consistency may be too difficult for the language models. Doing so may introduce too much noise and hurt translation quality.  

This method also provides an efficient way to assess a data augmentation scheme for NMT. It can save days or even months of GPU time (for training NMT models) since computing the word prediction accuracy rates on a few thousands of sentence pairs is very fast.

\subsection{Encoder vs. Decoder}
\label{sec.enc_dec}

Our CMLM-based data augmentation method can be applied to either encoder or decoder, or both. In this section, we conduct experiments to study the effects of these choices. We train two CMLMs independently. The first is used to augment the encoder, the latter the decoder. Note that, per our discussion in section \ref{sec.cmlm}, the CMLMs, when activated, only augment one side of the sentence pair. The encoder (resp. decoder) CMLM mask out words \textbf{only} in $X$ (resp. $Y$), thus replacing their embeddings by their soft versions.

Table \ref{table:2} shows the BLEU scores for different augmentation configurations. It is clear that both encoder and decoder augmentations are beneficial, with encoder augmentation obtaining slightly more gain. The maximum improvement can be achieved when the method is applied to both.


\begin{table}[h]
\begin{tabular}{lcccc}
\hline
\multicolumn{1}{c}{\multirow{2}{*}{}} & \multicolumn{3}{c}{\textbf{IWSLT}}                                                & \multicolumn{1}{c}{\textbf{WMT}}   \\ \cline{2-5} 
\multicolumn{1}{c}{}                  & \multicolumn{1}{c}{De-En} & \multicolumn{1}{c}{Es-En} & \multicolumn{1}{c}{He-En} & \multicolumn{1}{c}{En-De} \\ \hline
\multicolumn{1}{l}{Base}              & \multicolumn{1}{c}{34.37} & \multicolumn{1}{c}{41.67} & \multicolumn{1}{c}{33.76} & \multicolumn{1}{c}{28.25} \\ 
+Encoder                              & 35.23                     & 42.31                     & 34.66                     & 29.57                     \\ 
+Decoder                              & 34.93                     & 42.13                     & 34.41                     & 29.34                     \\ 
+Both                                 & 35.93                     & 42.92                     & 35.21                     & 30.15                     \\ \hline
\end{tabular}
\caption{BLEU scores over the test sets.}
\label{table:2}
\end{table}

\subsection{Mask Probability}
\label{sec.gamma}

\begin{figure}[h]
  \centering
  \includegraphics[width=0.5\textwidth]{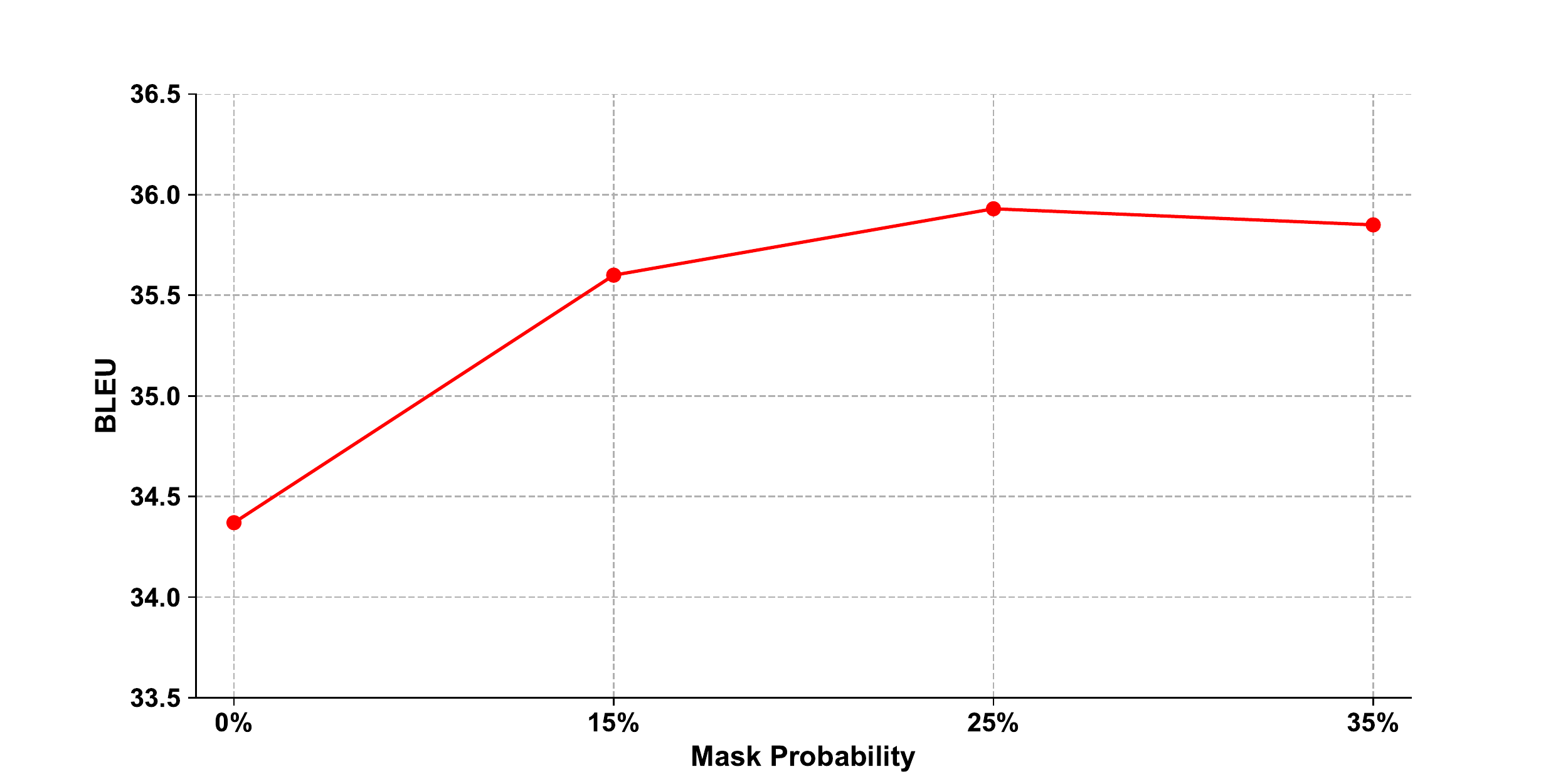}
  \caption{BLEU score for IWSLT14 German-English with difference mask probability.}
  \label{fig:mask_probability}
\end{figure}

As mentioned in section \ref{sec.soft}, for each word in $X$ or $Y$, we replace its embedding by its soft version with probability $\gamma$. This parameter controls the extent to which the DA method will exert its effect. Intuitively, a small value of $\gamma$ will preserve the original semantics better while a large value of $\gamma$ can bring in more diversity. A balance must be struck. We experiment with different values, and Figure \ref{fig:mask_probability} shows their influence on BLEU on the IWSLT14 De-En dataset. The strongest performance is reached with a mask probability of 0.25.

\subsection{Computation Overhead}

Our DA method introduces two additional steps into the NMT training process: fine-tuning the CMLMs and augmenting the NMT model. The actual overhead depends on the scale of the data sets. In our experiments, IWSLT De-En and WMT En-De corpora consist of 160K and 4.5M sentence pairs, respectively. Fine-tuning the CMLMs on the two corpora takes about 3 and 20 hours, respectively, on a \textit{single} A40 GPU. 

Our training process has the same complexity as that of \textit{SCA} \citep{gao2019soft} so they should have similar computation performance. From our experiments, the training time on IWSLT dataset increases about 84\%, up from 2.5 hours to 4.6 hours, again on a \textit{single} A40. The overhead is less significant for large corpora. The WMT tasks take 25\% more time to train, up from one day to roughly 32 hours on 4 A40 cards. We see only a 10\% increase in training time when we apply the DA method to our production NMT engine.


\section{Conclusion}

In this paper, we advocate performing semantically consistent data augmentation for neural machine translation and propose a scheme based on Conditional Masked Language Model and soft word substitution. We show that a deep, bi-directional CMLM is capable of enforcing semantic consistency by conditioning on \textit{both} source and target during data augmentation. Experiments demonstrate that the overall solution results in more realistic data augmentation and better translation quality. Our approach consistently achieves the best performance in comparison with strong and recent works and yields improvements of up to 1.90 BLEU points over baseline.


\bibliography{anthology,custom}
\bibliographystyle{acl_natbib}

\end{document}